\documentclass{article}

\PassOptionsToPackage{numbers, compress}{natbib}

\usepackage[final]{neurips_2019}

\usepackage[utf8]{inputenc} 
\usepackage[T1]{fontenc}    
\usepackage{hyperref}       
\usepackage{url}            
\usepackage{booktabs}       
\usepackage{amsfonts}       
\usepackage{nicefrac}       
\usepackage{microtype}      

\usepackage{subcaption}
\usepackage{boldline}
\usepackage{cuted}
\usepackage{capt-of}
\usepackage[font=small]{caption}
\usepackage{tikz}
\usepackage{graphicx}
\usepackage{amsmath}
\usepackage{amssymb}
\usepackage{acronym}
\usepackage{verbatim}
\usepackage{xspace}
\usepackage{enumitem}
\usepackage{makecell}
\usepackage{setspace}
\usepackage{multirow}
\usepackage{balance}
\usepackage[compact]{titlesec}
\usepackage{setspace}
\usepackage{titlesec}
\usepackage[export]{adjustbox}

\makeatletter
\DeclareRobustCommand\onedot{\futurelet\@let@token\@onedot}
\def\@onedot{\ifx\@let@token.\else.\null\fi\xspace}
\def\eg{\emph{e.g}\onedot} 
\def\ie{\emph{i.e}\onedot}

\makeatother

\title{PerspectiveNet: 3D Object Detection from\\a Single RGB Image via Perspective Points}

\author{%
  Siyuan Huang \\
  Department of Statistics\\
  \texttt{huangsiyuan@ucla.edu}\\
  \And
  Yixin Chen\\
  Department of Statistics\\
  \texttt{ethanchen@ucla.edu}\\
  \And
  Tao Yuan\\
  Department of Statistics\\
  \texttt{taoyuan@ucla.edu}\\
  \AND
  Siyuan Qi\\
  Department of Computer Science\\
  \texttt{syqi@cs.ucla.edu}\\
  \And
  Yixin Zhu\\
  Department of Statistics\\
  \texttt{yixin.zhu@ucla.edu}\\
  \And
  Song-Chun Zhu\\
  Department of Statistics\\
  \texttt{sczhu@stat.ucla.edu}\\
}

\begin{document}
\maketitle

\setstretch{0.98}

\begin{abstract}
Detecting 3D objects from a single RGB image is intrinsically ambiguous, thus requiring appropriate prior knowledge and intermediate representations as constraints to reduce the uncertainties and improve the consistencies between the 2D image plane and the 3D world coordinate. To address this challenge, we propose to adopt perspective points as a new intermediate representation for 3D object detection, defined as the 2D projections of local Manhattan 3D keypoints to locate an object; these perspective points satisfy geometric constraints imposed by the perspective projection. We further devise PerspectiveNet, an end-to-end trainable model that simultaneously detects the 2D bounding box, 2D perspective points, and 3D object bounding box for each object from a single RGB image. PerspectiveNet yields three unique advantages: (i) 3D object bounding boxes are estimated based on perspective points, bridging the gap between 2D and 3D bounding boxes \emph{without} the need of category-specific 3D shape priors. (ii) It predicts the perspective points by a \emph{template-based} method, and a perspective loss is formulated to maintain the perspective constraints. (iii) It maintains the consistency between the 2D perspective points and 3D bounding boxes via a \emph{differentiable} projective function. Experiments on SUN RGB-D dataset show that the proposed method significantly outperforms existing RGB-based approaches for 3D object detection.
\end{abstract}

\section{Introduction}
\label{sec:intro}

\begin{quote}
    If one hopes to achieve a full understanding of a system as complicated as a nervous system, \ldots, or even a large computer program, then one must be prepared to contemplate different kinds of explanation at different levels of description that are linked, at least in principle, into a cohesive whole, even if linking the levels in complete details is impractical. \hfill---~David Marr~\cite{marr1982vision}, pp. 20--21
\end{quote}

In a classic view of computer vision, David Marr~\cite{marr1982vision} conjectured that the perception of a 2D image is an \emph{explicit} multi-phase information process, involving (i) an early vision system of perceiving textures~\cite{julesz1962visual,zhu1998filters} and textons~\cite{julesz1981textons,zhu2005textons} to form a primal sketch as a perceptually lossless conversion from the raw image~\cite{guo2003towards,guo2007primal}, (ii) a mid-level vision system to construct 2.1D (multiple layers with partial occlusion)~\cite{nitzberg19902,wang1993layered,wang1994representing} and 2.5D~\cite{marr1978representation} sketches, and (iii) a high-level vision system that recovers the full 3D~\cite{binford1971visual,brooks1981symbolic,kanade1981recovery}. In particular, he highlighted the importance of different levels of organization and the internal representation~\cite{broadbent1985question}.

In parallel, the school of Gestalt Laws~\citep{wertheimer1912experimentelle,wagemans2012century,wagemans2012century2,kohler1920physischen,kohler1938physical,wertheimer1923untersuchungen,wertheimer1938laws,koffka2013principles} and perceptual organization~\cite{lowe2012perceptual,pentland1987perceptual} aims to resolve the 3D reconstruction problem from a single RGB image without forming the depth cues; but rather, they often use some sorts of priors---groupings and structural cues~\cite{waltz1975understanding,barrow1981interpreting} that are likely to be invariant over wide ranges of viewpoints~\cite{lowe1987three}, resulting in the birth of the SIFT feature~\cite{lowe2004distinctive}. Later, from a Bayesian perspective at a scene level, such priors, independent of any 3D scene structures, were found in the human-made scenes, known as the Manhattan World assumption~\cite{coughlan2003manhattan}. Importantly, further studies found that such priors help to improve object detection~\cite{coughlan1999manhattan}.

In this paper, inspired by these two classic schools in computer vision, we seek to test the following two hypotheses using modern computer vision methods: (i) Could an \emph{intermediate representation} facilitate modern computer vision tasks? (ii) Is such an intermediate representation a better and more \emph{invariant prior} compared to the priors obtained directly from specific tasks?

In particular, we tackle the challenging task of 3D object detection from a single RGB image. Despite the recent success in 2D scene understanding (\eg,~\cite{ren2015faster,he2017mask}, there is still a significant performance gap for 3D computer vision tasks based on a single 2D image. Recent modern approaches directly regress the 3D bounding boxes~\cite{chen2016monocular,mousavian20173d,huang2018cooperative} or reconstruct the 3D objects with specific 3D object priors~\cite{kundu20183d,huang2018holistic,yao20183d,he2019mono3d++}. In contrast, we propose an end-to-end trainable framework, PerspectiveNet, that sequentially estimates the 2D bounding box, 2D perspective points, and 3D bounding box for each object with a local Manhattan assumption~\cite{xiao2014reconstructing}, in which the perspective points serve as the intermediate representation, defined as the 2D projections of local Manhattan 3D keypoints to locate an object.

\begin{figure}[t!]
    \begin{center}
        \includegraphics[width=\linewidth]{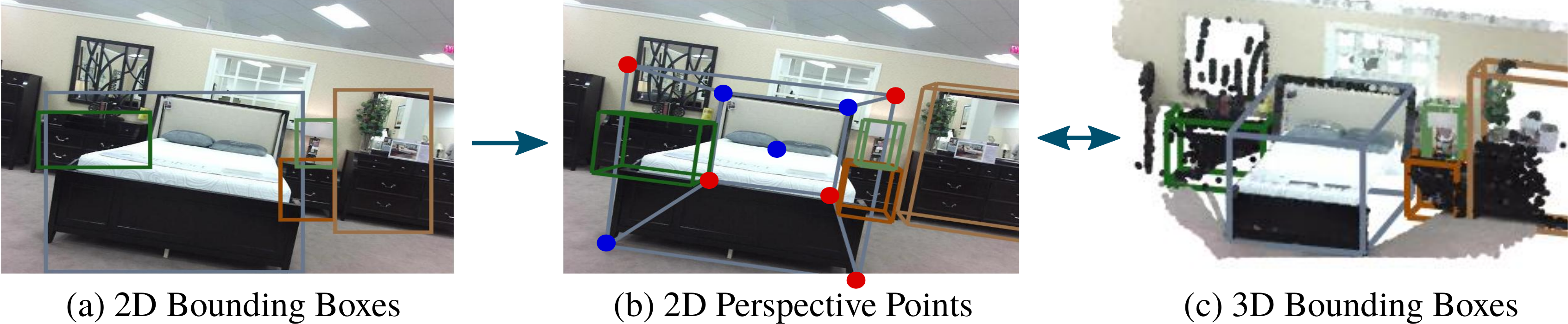}
    \end{center}
    \caption{Traditional 3D object detection methods directly estimate (c) the 3D object bounding boxes from (a) the 2D bounding boxes, which suffer from the uncertainties between the 2D image plane and the 3D world. The proposed PerspectiveNet utilizes (b) the 2D perspective points as the intermediate representation to bridge the gap. The perspective points are the 2D perspective projection of the 3D bounding box corners, containing rich 3D information (\eg, positions, orientations). The red dots indicate the perspective points of the bed that are challenging to emerge based on the visual features, but could be inferred by the context (correlations and topology) among other perspective points.}
    \label{fig:overview}
\end{figure}

The proposed method offers three unique advantages. First, the use of perspective points as the \emph{intermediate representation} bridges the gap between 2D and 3D bounding boxes \emph{without} utilizing any extra category-specific 3D shape priors. As shown in \autoref{fig:overview}, it is often challenging for learning-based methods to estimate the 3D bounding boxes from 2D images directly; regressing 3D bounding boxes from 2D input is a highly under-constrained problem and can be easily influenced by appearance variations of shape, texture, lighting, and background. To alleviate this issue, we adopt the perspective points as an intermediate representation to represent the local Manhattan frame that each 3D object aligns with. Intuitively, the perspective points of an object are \emph{3D geometric constraints in the 2D space}. More specifically, the 2D perspective points for each object are defined as the perspective projection of the 3D object bounding box (concatenated with its center), and each 3D box aligns within a 3D local Manhattan frame. These perspective points are fused into the 3D branch to predict the 3D attributes of the 3D bounding boxes.

Second, we devise a \emph{template-based} method to efficiently and robustly estimate the perspective points. Existing methods~\cite{newell2016stacked,lee2017roomnet,zou2018layoutnet,he2017mask,suwajanakorn2018discovery} usually exploit heatmap or probability distribution map as the representation to learn the location of visual points (\eg, object keypoint, human skeleton, room layout), relying heavily on the view-dependent visual features, thus insufficient to resolve occlusions or large rotation/viewpoint changes in complex scenes; see an example in \autoref{fig:overview} (b) where the five perspective points (in red) are challenging to emerge from pure visual features but could be inferred by the correlations and topology among other perspective points. To tackle this problem, we treat each set of 2D perspective points as the low dimensional embedding of its corresponding set of 3D points with a constant topology; such an embedding is learned by predicting the perspective points as a mixture of sparse templates. A perspective loss is formulated to impose the perspective constraints; the details are described in \autoref{sec:perspective}.

Third, the consistency between the 2D perspective points and 3D bounding boxes can be maintained by a \emph{differentiable} projective function; it is end-to-end trainable, from the 2D region proposals, to the 2D bounding boxes, to the 2D perspective points, and to the 3D bounding boxes.

In the experiment, we show that the proposed PerspectiveNet outperforms previous methods with a large margin on SUN RGB-D dataset~\cite{song2015sun}, demonstrating its efficacy on 3D object detection.

\setstretch{0.97}

\section{Related Work}

\paragraph{3D object detection from a single image}

Detecting 3D objects from a single RGB image is a challenging problem, particularly due to the intrinsic ambiguity of the problem. Existing methods could be categorized into three streams: (i) geometry-based methods that estimate the 3D bounding boxes with geometry and 3D world priors~\cite{zhao2011image,zhao2013scene,choi2013understanding,lin2013holistic,zhang2014panocontext}; (ii) learning-based methods that incorporate category-specific 3D shape prior~\cite{izadinia2016im2cad,huang2018holistic,he2019mono3d++} or extra 2.5D information (depth, surface normal, and segmentation)~\cite{kundu20183d,yao20183d,xu2018multi} to detect 3D bounding boxes or reconstruct the 3D object shape; and (iii) deep learning methods that directly estimates the 3D object bounding boxes from 2D bounding boxes~\cite{chen20153d,chen2016monocular,mousavian20173d,huang2018cooperative}. To make better estimations, various techniques have been devised to enforce consistencies between the estimated 3D and the input 2D image. \citet{huang2018cooperative} proposed a two-stage method to learn the 3D objects and 3D layout cooperatively. \citet{kundu20183d} proposed a 3D object detection and reconstruction method using category-specific object shape prior by render-and-compare. Different from these methods, the proposed PerspectiveNet is a one-stage end-to-end trainable 3D object detection framework using perspective points as an intermediate representation; the perspective points naturally bridge the gap between the 2D and 3D bounding boxes without any extra annotations, category-specific 3D shape priors, or 2.5D maps.

\paragraph{Manhattan World assumption}

Human-made environment, from the layout of a city to structures such as buildings, room, furniture, and many other objects, could be viewed as a set of parallel and orthogonal planes, known as the Manhattan World (MW) assumption~\cite{coughlan1999manhattan}. Formally, it indicates that most human-made structures could be approximated by planar surfaces that are parallel to one of the three principal planes of a common orthogonal coordinate system. This strict Manhattan World assumption is later extended by a Mixture of Manhattan Frame (MMF)~\cite{straub2014mixture} to represent more complex real-world scenes (\eg, city layouts, rotated objects). In literature, MW and MMF have been adopted in vanish points (VPs) estimation and camera calibration~\cite{schindler2004atlanta,kroeger2015joint}, orientation estimation~\cite{bosse2003vanishing,straub2015real,ghanem2015robust}, layout estimation~\cite{hedau2009recovering,lee2009geometric,hedau2010thinking,schwing2012efficient,zou2018layoutnet}, and 3D scene reconstruction~\cite{delage2007automatic,furukawa2009manhattan,xiao2013basic,xiao2014reconstructing,ren2016three,liu2017single}. In this work, we extend the MW to local Manhattan assumption where the cuboids are aligned with the vertical (gravity) direction but with arbitrary horizontal orientation (also see \citet{xiao2014reconstructing}), and perspective points are adopted as the intermediate representation for 3D object detection.

\paragraph{Intermediate 3D representation}

Intermediate 3D representations are bridges that narrow the gap and maintain the consistency between the 2D image plane and 3D world. Among them, 2.5D sketches have been broadly used in reconstructing the 3D shapes~\cite{wu2017marrnet,von2018,genre2018} and 3D scenes~\cite{tulsiani2017factoring,huang2018holistic}. Other recent alternative intermediate 3D representations include: (i) \citet{wu2016single} uses pre-annotated and category-specific object keypoints as an intermediate representation, and (ii) \citet{tekin2018real} uses projected corners of 3D bounding boxes in learning the 6D object pose. In this paper, we explore the perspective points as an intermediate representation of 2D and 3D bounding boxes, and provide an efficient learning framework for 3D object detection.

\section{Learning Perspective Points for 3D Object Detection}

\subsection{Overall Architecture}

\begin{figure}[t!]
    \begin{center}
        \includegraphics[width=\linewidth]{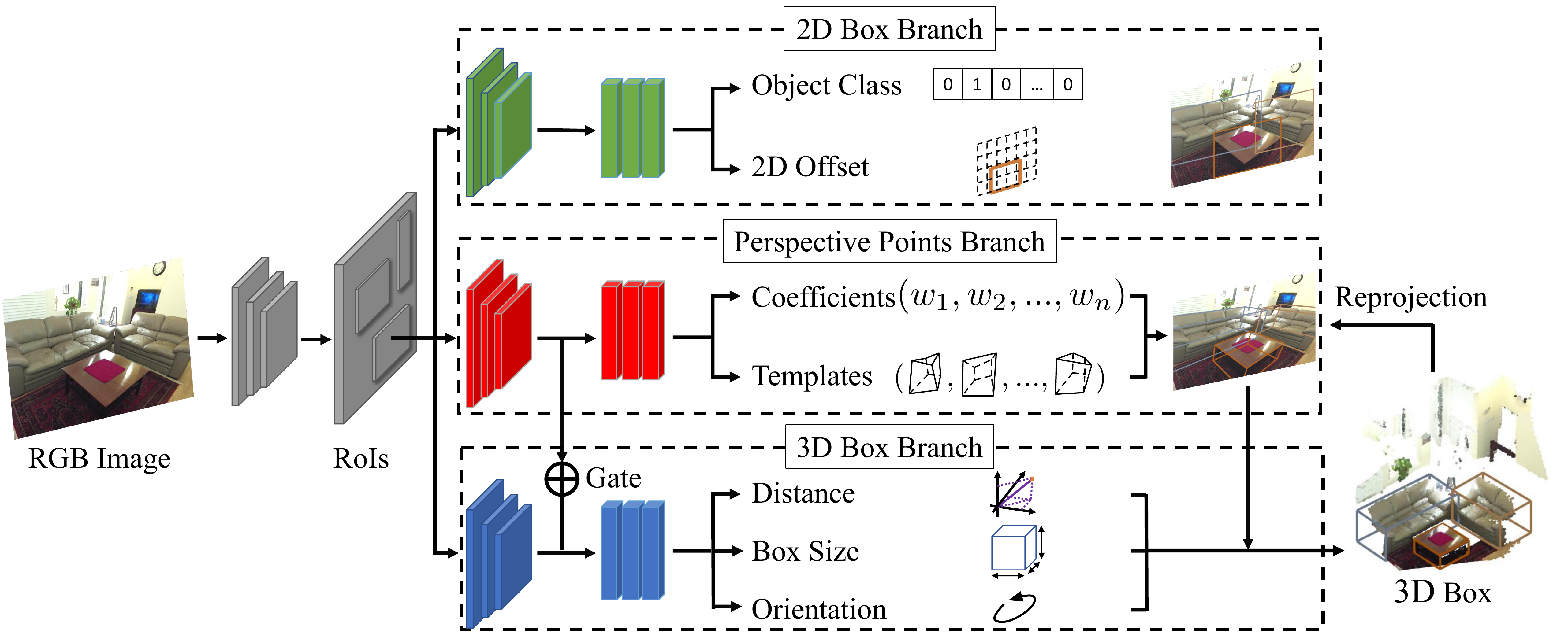}
    \end{center}
    \vspace{-6pt}
    \caption{The proposed framework of the PerspectiveNet. Given an RGB Image, the backbone of PerspectiveNet extracts global features and propose candidate 2D bounding boxes (RoIs). For each proposed box, its RoI feature is fed into three network branches to predict: (i) the object class and the 2D box offset, (ii) 2D perspective templates (projected 3D box corners and object center) and the corresponding coefficients, and (iii) the 3D box size, orientation, and its distance from the camera. Detected 3D boxes are reconstructed by the projected object center, distance, box size, and rotation. By projecting the detected 3D boxes to 2D and comparing them with 2D perspective points, the network imposes and learns a consistency between the 2D inputs and 3D estimations.}
    \label{fig:framework}
    \vspace{6pt}
\end{figure}

As shown in \autoref{fig:framework}, the proposed PerspectiveNet contains a backbone architecture for feature extraction over the entire image, a region proposal network (RPN)~\cite{ren2015faster} that proposes regions of interest (RoIs), and a network head including three region-wise parallel branches.
For each proposed box, its RoI feature is fed into the three network branches to predict: (i) the object class and the 2D bounding box offset, (ii) the 2D perspective points (projected 3D box corners and object center) as a weighted sum of predicted perspective templates, and (iii) the 3D box size, orientation, and its distance from the camera. Detected 3D boxes are reconstructed by the projected object center, distance, box size, and rotation.
The overall architecture of the PerspectiveNet resembles the R-CNN structure, and we refer readers to~\cite{ren2015faster,girshick2015fast,he2017mask} for more details of training R-CNN detectors.

During training, we define a multi-task loss on each proposed RoI as
\begin{equation}
    \mathcal{L} = \mathcal{L}_{cls} + \mathcal{L}_{2D} + \mathcal{L}_{pp} + \mathcal{L}_{p} + \mathcal{L}_{3D} + \mathcal{L}_{proj},
\end{equation}
where the classification loss $\mathcal{L}_{cls}$ and 2D bounding box loss $\mathcal{L}_{2D}$ belong to the 2D bounding box branch and are identical to those defined in 2D R-CNNs~\cite{ren2015faster,he2017mask}. $\mathcal{L}_{pp}$ and $\mathcal{L}_{p}$ are defined on the perspective point branch (\autoref{sec:perspective}), $\mathcal{L}_{3D}$ is defined on the 3D bounding box branch (see \autoref{sec:3dbox}), and the $\mathcal{L}_{proj}$ is defined on maintaining the 2D-3D projection consistency (see \autoref{sec:consistency}).

\setstretch{0.98}

\subsection{Perspective Point Estimation}
\label{sec:perspective}

The perspective point branch estimates the set of 2D perspective points for each RoI. Formally, the 2D perspective points of an object are the 2D projections of local Manhattan 3D keypoints to locate that object, and they satisfy certain geometric constraints imposed by the perspective projection. In our case, the perspective points (\autoref{fig:overview}(b)) include the 2D projections of the 3D bounding box corners and the 3D object center. The perspective points are predicted using a template-based regression and learned by a mean squared error and a perspective loss detailed below.

\subsubsection{Template-based Regression}

Most of the existing methods~\cite{newell2016stacked,lee2017roomnet,zou2018layoutnet,he2017mask,suwajanakorn2018discovery} estimate visual keypoints with heatmaps, where each map predicts the location for a certain keypoint. However, predicting perspective points by heatmaps has two major problems: (i) Heatmap prediction for different keypoints is independent, thus fail to capture the topology correlations among the perspective points. (ii) Heatmap prediction for each keypoint relies heavily on the visual feature such as corners, which may be difficult to detect (see an example in \autoref{fig:overview}(b)). In contrast, each set of 2D perspective points can be treated as a low dimensional embedding of a set of 3D points with a particular topology, thus inferring such points relies more on the relation and topology among the points instead of just the visual features.

To tackle these problems, we avoid dense per-pixel predictions. Instead, we estimate the perspective points by a mixture of sparse templates~\cite{olshausen1996emergence,wu2010learning}. The sparse templates are more robust when facing unfamiliar scenes or objects. Ablative experiments show that the proposed template-based method provides a more accurate estimation of perspective points than heatmap-based methods; see \autoref{sec:ablative}.

Specifically, we project both the 3D object center and eight 3D bounding box corners to 2D with camera parameters to generate the ground-truth 2D perspective points $P_{gt} \in \mathbb{R}^{2 \times 9}$. Since a portion of the perspective points usually lies out of the RoI, we calculate the location of the perspective points in an extended (doubled) size of RoI and normalize the locations to $[0, 1]$.

\begin{figure}[t!]
    \begin{center}
        \includegraphics[width=\linewidth]{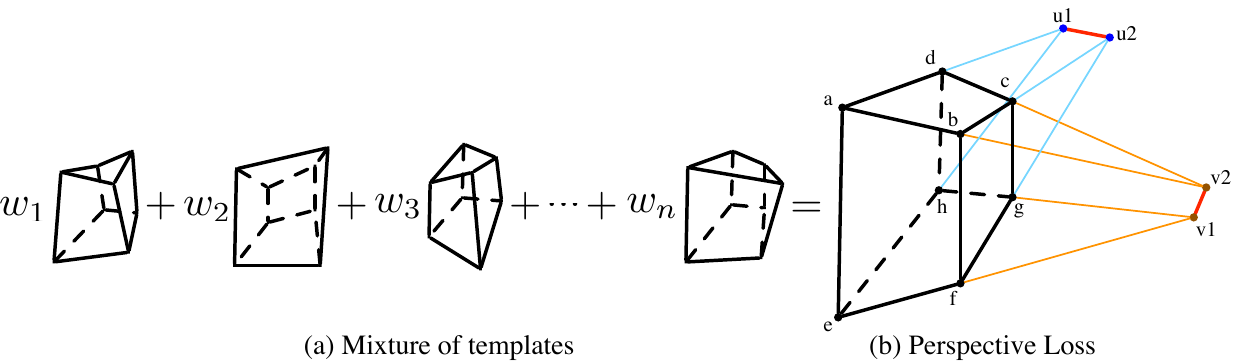}
    \end{center}
    \caption{Perspective point estimation. (a) The perspective points are estimated by a mixture of templates through a linear combination. Each template encodes geometric cues including orientations and viewpoints. (b) The perspective loss enforces each set of 2D perspective points to be the perspective projection of a (vertical) 3D cuboid. For a vertical cuboid, the projected vertical edges (\ie, $ae$, $bf$, $cg$, and $dh$) should be parallel or near parallel (under small camera tilting angles). For 3D parallel lines that are perpendicular to the gravity direction, the vanishing points of their 2D projections should coincide (\eg, $u1$ and $u2$).}
    \label{fig:perspective}
\end{figure}

We predict the perspective points by a linear combination of templates; see \autoref{fig:perspective}. The perspective point branch has a $C \times K \times 2 \times 9$ dimensional output for the templates $T$, and a $C \times K$ dimensional output for the coefficients $w$, where $K$ denotes the number of templates for each class and $C$ denotes the number of object classes. The templates $T$ is scaled to $[0, 1]$ by a sigmoid nonlinear function, and the coefficients $w$ is normalized by a softmax function. The estimated perspective points $\hat{P} \in \mathbb{R}^{C \times 2 \times 9}$ can be computed by a linear combination: 
\begin{equation}
    \hat{P_i} = \sum^{K}_{k=1} w_{ik}\,T_{ik}, \quad{} \forall i = 1, \cdots, C.
\end{equation}

\setstretch{1}

The template design is both class-specific and instance-specific: (i) Class-specific: we decouple the prediction of the perspective point and the object class, allowing the network to learn perspective points for every class without competition among classes. (ii) Instance-specific: the templates are inferred for each RoI; hence, they are specific to each object instance. The templates are automatically learned for each object instance from data with the end-to-end learning framework; thus, both the templates and coefficients for each instance are optimizable and can better fit the training data.

The average mean squared error (MSE) loss is defined as $\mathcal{L}_{pp} = \text{MSE} (\hat{P}_c, P_{gt})$. For an RoI associated with ground-truth class $c$, $\mathcal{L}_{pp}$ is only defined on the $c$'s perspective points during training; perspective point outputs from other classes do not contribute to the loss. In inference, we rely on the dedicated classification branch to predict the class label to select the output perspective points.

\subsubsection{Perspective Loss}

Under the assumption that each 3D bounding box aligns with a local Manhattan frame, we regularize the estimation of the perspective points to satisfy the constraint of perspective projection. Each set of mutually parallel lines in 3D can be projected into 2D as intersecting lines; see \autoref{fig:perspective} (b). These intersecting lines should converge at the same vanishing point. Therefore, the desired algorithm would penalize the distance between the intersection points from the two sets of intersecting lines. For example in \autoref{fig:perspective} (b), we select line $ad$ and line $eh$ as a pair of lines, $bc$ and $fg$ as another, and compute the distance between their intersection point $u_1$ and $u_2$. Additionally, since we assume each 3D local Manhattan frame aligns with the vertical (gravity) direction, we enforce the edges in gravity direction (\ie, $ae$, $bf$, $cg$, and $dh$) to be parallel by penalizing the large slope variance.

The perspective loss is computed as $\mathcal{L}_{p} = \mathcal{L}_{d1} + \mathcal{L}_{d2} + \mathcal{L}_{grav}$, where $\mathcal{L}_{grav}$ penalizes the slope variance in gravity direction, $\mathcal{L}_{d1}$ and $\mathcal{L}_{d2}$ penalize the intersection point distance for the two perpendicular directions along the gravity direction.

\subsection{3D Bounding Box Estimation}
\label{sec:3dbox}

Estimating 3D bounding boxes is a two-step process. In the first step, the 3D branch estimates the 3D attributes, including the distance between the camera center and the 3D object center, as well as the 3D size and orientation following \citet{huang2018cooperative}. Since the perspective point branch encodes rich 3D geometric features, the 3D attribute estimator aggregates the feature from perspective point branch with a soft gated function between $[0, 1]$ to improve the prediction. The gated function serves as a soft-attention mechanism that decides how much information from perspective points should contribute to the 3D prediction.

In the second step, with the estimated projected 3D bounding boxes center (\ie, the first estimated perspective point) and the 3D attributes, we compose the 3D bounding boxes by the inverse projection from the 2D image plane to the 3D world following \citet{huang2018cooperative} given camera parameters.

The 3D loss is computed by the sum of individual losses of 3D attributes and a joint loss of 3D bounding box $\mathcal{L}_{3D} = \mathcal{L}_{dis} + \mathcal{L}_{size} + \mathcal{L}_{ori} + \mathcal{L}_{box3d}$.

\setstretch{1}

\subsection{2D-3D Consistency}
\label{sec:consistency}

In contrast to prior work~\cite{wu2016single,rezende2016unsupervised,yan2016perspective,mousavian20173d,wu2017marrnet,huang2018cooperative} that enforces the consistency between estimated 3D objects and 2D image, we devise a new way to impose a re-projection consistency loss between 3D bounding boxes and perspective points. Specifically, we compute the 2D projected perspective points $P_{proj}$ by projecting the 3D bounding box corners back to 2D image plane and computing the distance with respect to ground-truth perspective points $\mathcal{L}_{proj} = \text{MSE} (P_{proj}, P_{gt})$. Comparing with prior work to maintain the consistency between 2D and 3D bounding boxes by approximating the 2D projection of 3D bounding boxes~\cite{mousavian20173d,huang2018cooperative}, the proposed method uses the \emph{exact} projection of projected 3D boxes to establish the consistency, capturing a more precise 2D-3D relationship.

\begin{figure}[t!]
    \begin{center}
        \includegraphics[width=\linewidth]{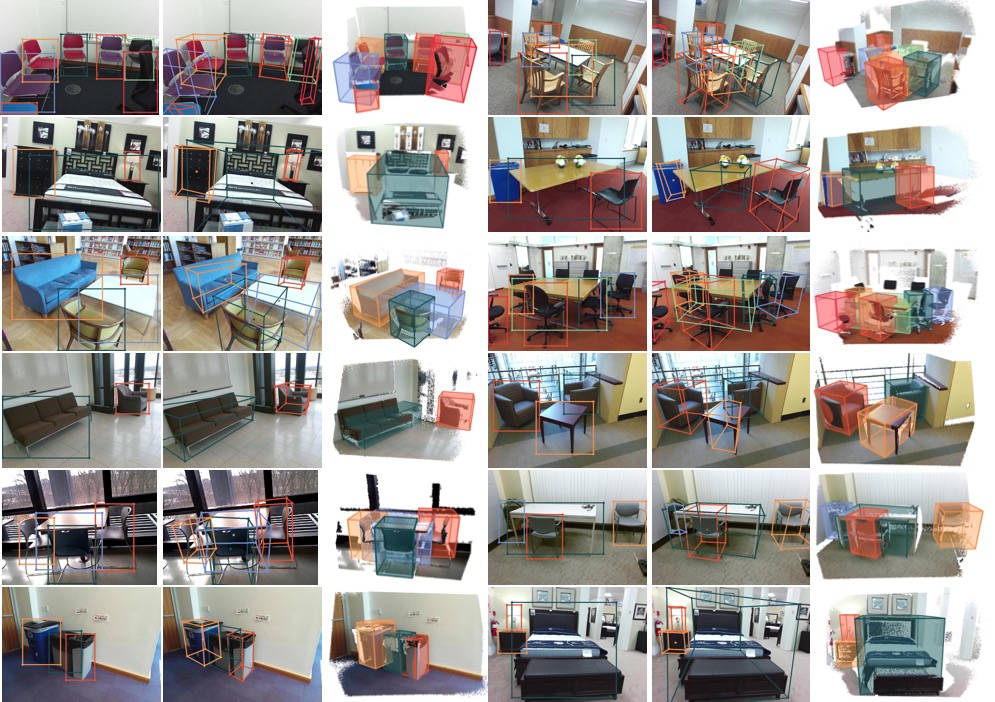}
    \end{center}
    \vspace{-6pt}
    \caption{Qualitative results (top 50\%). For every three columns as a group: (Left) The RGB image with 2D detection results. (Middle) The RGB image with estimated perspective points. (Right) The results in 3D point cloud; point cloud is used for visualization only.}
    \label{fig:result}
    \vspace{6pt}
\end{figure}

\section{Implementation Details}
\label{sec:details}

\paragraph{Network Backbone} Inspired by \citet{he2017mask}, we use the combination of residual network (ResNet)~\cite{he2016deep} and feature pyramid network (FPN)~\cite{lin2017feature} to extract the feature from the entire image. A region proposal network (RPN)~\cite{ren2015faster} is used to produce object proposals (\ie, RoI). A RoIAlign~\cite{he2017mask} module is adopted to extract a smaller features map ($256 \times 7\times7$) for each proposal.

\paragraph{Network Head} The network head consists of three branches, and each branch has its individual feature extractor and predictor. Three feature extractors have the same architecture of two fully connected (FC) layers; each FC layer is followed by a ReLU function. The feature extractors take the $256 \times 7 \times 7$ dimensional RoI features as the input and output a 1024 dimensional vector.

The predictor in the 2D branch has two separate FC layers to predict a $C$ dimensional object class probabilities and a $C \times 4$ dimensional 2D bounding box offset. The predictor in the perspective point branch predicts $C\times K \times 2 \times 9$ dimensional templates and $C \times K$ dimensional coefficients with two FC layers and their corresponding nonlinear activation functions (\ie, sigmoid, softmax). The soft gate in the 3D branch consists of an FC layer (1024-1) and a sigmoid function to generate the weight for feature aggregation. The predictor in the 3D branch consists of three FC layers to predict the size, the distance from the camera, and the orientation of the 3D bounding box.

\setstretch{1}

\begin{figure}[t!]
    \begin{center}
        \includegraphics[width=\linewidth]{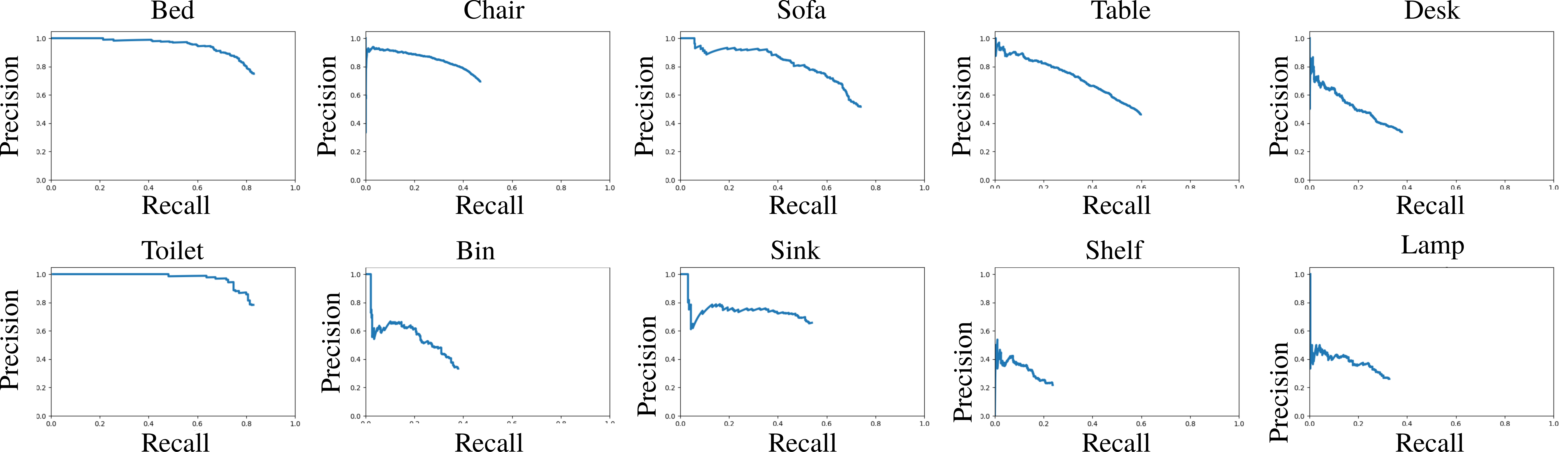}
    \end{center}
    \caption{Precision-Recall (PR) curves for 3D object detection on SUN RGB-D}
    \label{fig:pr_30}
\end{figure}

\section{Experiments}

\textbf{Dataset}\quad{} We conduct comprehensive experiments on SUN RGB-D~\cite{song2015sun} dataset. The SUN RGB-D dataset has a total of 10,335 images, in which 5,050 are test images. It has a rich annotation of scene categories, camera pose, and 3D bounding boxes. We evaluate the 3D object detection results of the proposed PerspectiveNet, make comparisons with the state-of-the-art methods, and further examine the contribution of each module in ablative experiments.

\textbf{Experimental Setup}\quad{} To prepare valid data for training the proposed model, we discard the images with no 3D objects or incorrect correspondence between 2D and 3D bounding boxes, resulting 4783 training images and 4220 test images. We detect 30 categories of objects following \citet{huang2018cooperative}.

\textbf{Reproduciblity Details}\quad{} During training, an RoI is considered positive if it has the IoU with a ground-truth box of at least 0.5. $\mathcal{L}_{pp}$, $\mathcal{L}_{p}$, $\mathcal{L}_{3D}$, and $\mathcal{L}_{proj}$ are only defined on positive RoIs. Each image has N sampled RoIs, where the ratio of positive to negative is 1:3 following the protocol presented in \citet{girshick2015fast}.

We resize the images so that the shorter edges are all 800 pixels. To avoid over-fitting, a data augmentation procedure is performed by randomly flipping the images or randomly shifting the 2D bounding boxes with corresponding labels during the training. We use SGD for optimization with a batch size of 32 on a desktop with 4 Nvidia TITAN RTX cards (8 images each card). The learning rate starts at 0.01 and decays by 0.1 at 30,000 and 35,000 iterations. We implement our framework based on the code of \citet{massa2018mrcnn}. It takes 6 hours to train, and the trained PerspectiveNet provides inference in real-time (20 FPS) using a single GPU.

Since the consistency loss and perspective loss can be substantial during the early stage of the training process, we add them to the joint loss when the learning rate decays twice. The hyper-parameter (\eg, the weights of losses, the architecture of network head) is tuned empirically by a local search.

\textbf{Evaluation Metric}\quad{} We evaluate the performance of 3D object detection using the metric presented in \citet{song2015sun}. Specifically, we first calculate the 3D Intersection over Union (IoU) between the predicted 3D bounding boxes and the ground-truth 3D bounding boxes, and then compute the mean average precision (mAP). Following \citet{huang2018cooperative}, we set the 3D IoU threshold as 0.15 in the absence of depth information. 

\textbf{Qualitative Results}\quad{} The qualitative results of 2D object detection, 2D perspective point estimation, and 3D object detection are shown in \autoref{fig:result}. Note that the proposed method performs accurate 3D object detection in some challenging scenes. For the perspective point estimation, even though some of the perspective points are not aligned with image features, the proposed method can still localize their positions robustly. 

\textbf{Quantitative Results}\quad{} Since the state-of-the-art method~\cite{huang2018cooperative} learns the camera extrinsic parameters jointly, we provide two protocals for evaluations for a fair comparison: (i) PerspectiveNet given ground-truth camera extrinsic parameter (\emph{full}), and (ii) PerspectiveNet without ground-truth camera extrinsic parameter by learning it jointly following~\cite{huang2018cooperative} (\emph{w/o.~cam}).

We learn the detector for 30 object categories and report the precision-recall (PR) curve of 10 main categories in \autoref{fig:pr_30}. We calculate the area under the curve to compute AP; \autoref{tab:detection_img} shows the comparisons of APs of the proposed models with existing approaches (see supplementary materials for the APs of all 30 categories).

\setstretch{0.93}

Note that the critical difference between the proposed model and the state-of-the-art method~\cite{huang2018cooperative} is the intermediate representation to learn the 2D-3D consistency. \citet{huang2018cooperative} uses 2D bounding boxes to enforce a 2D-3D consistency by minimizing the differences between projected 3D boxes and detected 2D boxes. In contrast, the proposed intermediate representation has a clear advantage since projected 3D boxes often are not 2D rectangles, and perspective points eliminate such errors.

Quantitatively, our full model improves the mAP of the state-of-the-art method~\cite{huang2018cooperative} by 14.71\%, and the model without the camera extrinsic parameter improves by 10.91\%. The significant improvement of the mAP demonstrates the efficacy of the proposed intermediate representation. We defer more analysis on how each component contributes to the overall performance in \autoref{sec:ablative}.

\begin{table}[b!]
    \caption{Comparisons of 3D object detection on SUN RGB-D (AP).}
    \setlength{\tabcolsep}{4pt}
    \centering
    \resizebox{\textwidth}{!}{
    \begin{tabular}{l c c c c c c c c c c c}
        \Xhline{2\arrayrulewidth}
         & bed & chair & sofa & table & desk & toilet  & bin  & sink & shelf & lamp & mAP\\
        \hline
        3DGP~\cite{choi2013understanding} & 5.62 & 2.31 & 3.24 & 1.23 & - & -  & - & - & - & - & -  \\
        HoPR~\cite{huang2018holistic} & 58.29 & 13.56 & 28.37 & 12.12 & 4.79 & 16.50  & 0.63 & 2.18 & 1.29 & 2.41 & 14.01  \\
        CooP~\cite{huang2018cooperative} & 63.58 & 17.12 & 41.22 & 26.21  &  9.55  & 58.55 & 10.19 & 5.34 & 3.01 & 1.75  & 23.65\\
        \hline
        Ours (w/o. cam) & 71.39 & 34.94 & 55.63 & 34.10 & 14.23 & 73.73 & 17.47 & 34.41 & 4.21 & 9.54 & 34.96\\
        Ours (full) & \textbf{79.69} & \textbf{40.42} & \textbf{62.35} & \textbf{44.12} & \textbf{20.19} & \textbf{81.22} & \textbf{22.42} & \textbf{41.35} & \textbf{8.29} & \textbf{13.14} & \textbf{39.09}\\
        \Xhline{2\arrayrulewidth}
    \end{tabular}}
    \label{tab:detection_img}
\end{table}

\subsection{Ablative Analysis}
\label{sec:ablative}

\begin{figure}[t!]
    \begin{center}
        \includegraphics[width=\linewidth]{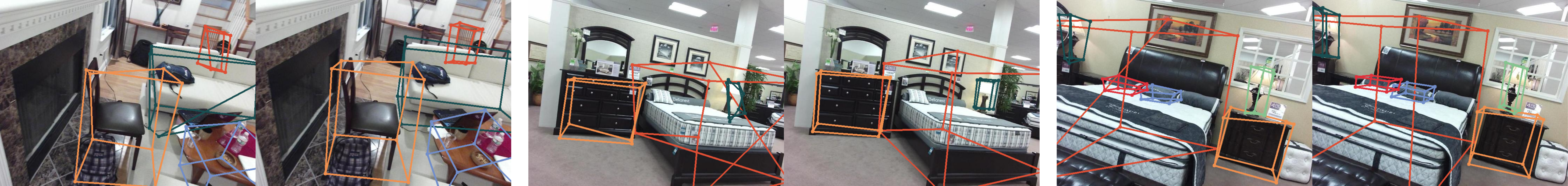}
    \end{center}
    \vspace{-6pt}
    \caption{Heatmaps vs. templates for perspective point prediction. (Left) Estimated by heatmap-based method. (Right) Estimated by the proposed template-based method.}
    \label{fig:heatmap}
\end{figure}

In this section, we analyze each major component of the model to examine its contribution to the overall significant performance gain. Specifically, we design six variants of the proposed model.\\
$\bullet$ $S_1$: The model trained without the perspective point branch, using the 2D offset to predict the 3D center of the object following \citet{huang2018cooperative}. \\
$\bullet$ $S_2$: The model that aggregates the feature from the perspective point branch and 3D branch directly without the gate function. \\
$\bullet$ $S_3$: The model that aggregates the feature from the perspective point branch and 3D branch with a gate function that only outputs 0 or 1 (hard gate). \\
$\bullet$ $S_4$: The model trained without the perspective loss. \\
$\bullet$ $S_5$: The model trained without the consistency loss. \\
$\bullet$ $S_6$: The model trained without the perspective branch, perspective loss, or consistency loss.\\
\autoref{tab:analysis} shows the mAP for each variant of the proposed model. The mAP drops 3.86\% without the perspective point branch ($S_1$), 1.66\% without the consistency loss ($S_5$), indicating that the perspective point and re-projection consistency influence the most to the proposed framework. In addition, the switch of gate function ($S_2$, $S_3$) and perspective loss ($S_4$) contribute less to the final performance. Since $S_6$ is still higher than the state-of-the-art result~\cite{huang2018cooperative} with 9.32\%, we conjecture this performance gain may come from the one-stage (vs. two-stage) end-to-end training framework and the usage of ground-truth camera parameter; we will further investigate this in future work.

\begin{table}[b!]
    \caption{Ablative analysis of the proposed model on SUN RGB-D. We evaluate the mAP for 3D object detection.}
    \setlength{\tabcolsep}{16pt}
    \centering
    \resizebox{\textwidth}{!}{
    \begin{tabular}{l|c|c|c|c|c|c|c}
        \Xhline{2\arrayrulewidth}
        \hline
        Setting & $S_1$ & $S_2$ & $S_3$ & $S_4$ & $S_5$ & $S_6$ & Full \\
        \hline
        mAP & 35.23 & 38.63 & 38.87 & 39.01 & 37.43 & 32.97 & \textbf{39.09} \\ 
        \Xhline{2\arrayrulewidth}
    \end{tabular}}
    \label{tab:analysis}
\end{table}

\subsection{Heatmaps vs. Templates}

As discussed in \autoref{sec:perspective}, we test two different methods for the perspective point estimation: (i) dense prediction as heatmaps following the human pose estimation mechanism in \citet{he2017mask} by adding a parallel heatmap prediction branch, and (ii) template-based regression by the proposed method. The qualitative results (see \autoref{fig:heatmap}) show that the heatmap-based estimation suffers severely from occlusion and topology change among the perspective points, whereas the proposed template-based regression eases the problem significantly by learning robust sparse templates, capturing consistent topological relations. We also evaluate the quantitative results by computing the average absolute distance between the ground-truth and estimated perspective points. The heatmap-based method has a 10.25 pix error, while the proposed method only has a 6.37 pix error, which further demonstrates the efficacy of the proposed template-based perspective point estimation.

\setstretch{1}

\subsection{Failure Cases}

\begin{figure}[t!]
    \begin{center}
        \includegraphics[width=\linewidth]{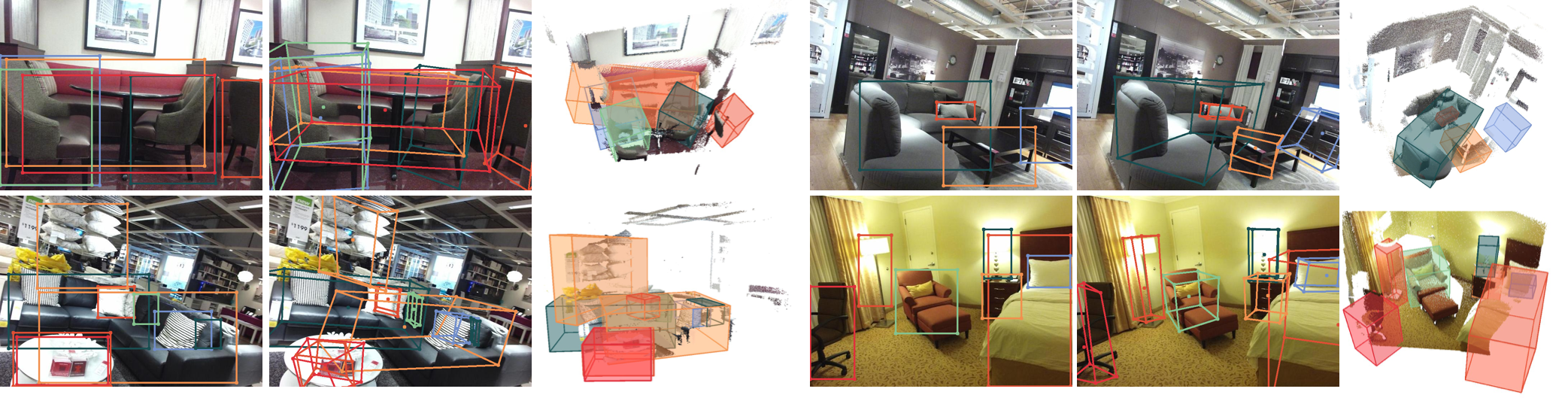}
    \end{center}
    \vspace{-3pt}
    \caption{Some failure cases. The perspective point estimation and the 3D box estimation fail at the same time.}
    \label{fig:result_fail}
    \vspace{9pt}
\end{figure}

In a large portion of the failure cases, the perspective point estimation and the 3D box estimation fail at the same time; see \autoref{fig:result_fail}. It implies that the perspective point estimation and the 3D box estimation are highly coupled, which supports the assumptions that the perspective points encode richer 3D information, and the 3D branch learns meaningful knowledge from the 2D branch. In future work, we may need a more sophisticated and general 3D prior to infer the 3D locations of objects for such challenging cases.

\subsection{Discussions and Future Work}

\paragraph{Comparison with optimization-based methods.}

Assume the estimated 3D size or distance is given, it is possible to compute the 3D bounding box with an optimization-based method like efficient PnP. However, the optimization-based methods are sensitive to the accuracy of the given known variables. It is more suitable for tasks with smaller solution spaces (\eg, 6-DoF pose estimation where the 3D shapes of objects are fixed). However, it would be difficult for tasks with larger solution spaces (\eg, 3D object detection where the 3D size, distance, and object pose could vary significantly). Therefore, we argue that directly estimating each variable with constraints imposed among them is a more natural and more straightforward solution.

\paragraph{Potential incorporation with depth information.}

The PerspectiveNet estimates the distance between the 3D object center and camera center based on the color image only (pure RGB without any depth information). If the depth information was also provided, the proposed method should be able to make a much more accurate distance prediction.

\paragraph{Potential application to outdoor environment.}

It would be interesting to see how the proposed method would perform on outdoor 3D object detection datasets like KITTI~\cite{geiger2013vision}. The differences between the indoor and outdoor datasets for the task of 3D object detection lie in various aspects, including the diversity of object categories, the variety of object dimension, the severeness of the occlusion, the range of the camera angles, and the range of the distance (depth). We hope to adopt the PerspectiveNet in future to the outdoor scenarios.

\section{Conclusion}

We propose the PerspectiveNet, an end-to-end differentiable framework for 3D object detection from a single RGB image. It uses perspective points as an intermediate representation between 2D input and 3D estimations. The PerspectiveNet adopts an R-CNN structure, where the region-wise branches predict 2D boxes, perspective points, and 3D boxes. Instead of using a direct regression of 2D-3D relations, we further propose a template-based regression for estimating the perspective points, which enforces a better consistency between the predicted 3D boxes and the 2D image input. The experiments show that the proposed method significantly improves existing RGB-based methods.

\paragraph{Acknowledgments}

This work reported herein is supported by MURI ONR N00014-16-1-2007, DARPA XAI N66001-17-2-4029, ONR N00014-19-1-2153, and an NVIDIA GPU donation grant.

\newpage
\setstretch{1}
\bibliographystyle{unsrtnat}
\setlength{\bibsep}{4pt}
\bibliography{neurips_2019}

\begin{thebibliography}{84}
\providecommand{\natexlab}[1]{#1}
\providecommand{\url}[1]{\texttt{#1}}
\expandafter\ifx\csname urlstyle\endcsname\relax
  \providecommand{\doi}[1]{doi: #1}\else
  \providecommand{\doi}{doi: \begingroup \urlstyle{rm}\Url}\fi

\bibitem[Marr(1982)]{marr1982vision}
David Marr.
\newblock \emph{Vision: A computational investigation into the human
  representation and processing of visual information}.
\newblock WH Freeman, 1982.

\bibitem[Julesz(1962)]{julesz1962visual}
Bela Julesz.
\newblock Visual pattern discrimination.
\newblock \emph{IRE transactions on Information Theory}, 8\penalty0
  (2):\penalty0 84--92, 1962.

\bibitem[Zhu et~al.(1998)Zhu, Wu, and Mumford]{zhu1998filters}
Song~Chun Zhu, Yingnian Wu, and David Mumford.
\newblock Filters, random fields and maximum entropy (frame): Towards a unified
  theory for texture modeling.
\newblock \emph{International Journal of Computer Vision (IJCV)}, 27\penalty0
  (2):\penalty0 107--126, 1998.

\bibitem[Julesz(1981)]{julesz1981textons}
Bela Julesz.
\newblock Textons, the elements of texture perception, and their interactions.
\newblock \emph{Nature}, 290\penalty0 (5802):\penalty0 91, 1981.

\bibitem[Zhu et~al.(2005)Zhu, Guo, Wang, and Xu]{zhu2005textons}
Song-Chun Zhu, Cheng-En Guo, Yizhou Wang, and Zijian Xu.
\newblock What are textons?
\newblock \emph{International Journal of Computer Vision (IJCV)}, 62\penalty0
  (1-2):\penalty0 121--143, 2005.

\bibitem[Guo et~al.(2003)Guo, Zhu, and Wu]{guo2003towards}
Cheng-en Guo, Song-Chun Zhu, and Ying~Nian Wu.
\newblock Towards a mathematical theory of primal sketch and sketchability.
\newblock In \emph{International Conference on Computer Vision (ICCV)}, 2003.

\bibitem[Guo et~al.(2007)Guo, Zhu, and Wu]{guo2007primal}
Cheng-en Guo, Song-Chun Zhu, and Ying~Nian Wu.
\newblock Primal sketch: Integrating structure and texture.
\newblock \emph{Computer Vision and Image Understanding (CVIU)}, 106\penalty0
  (1):\penalty0 5--19, 2007.

\bibitem[Nitzberg and Mumford(1990)]{nitzberg19902}
Mark Nitzberg and David Mumford.
\newblock The 2.1-d sketch.
\newblock In \emph{ICCV}, 1990.

\bibitem[Wang and Adelson(1993)]{wang1993layered}
John~YA Wang and Edward~H Adelson.
\newblock Layered representation for motion analysis.
\newblock In \emph{Conference on Computer Vision and Pattern Recognition
  (CVPR)}, 1993.

\bibitem[Wang and Adelson(1994)]{wang1994representing}
John~YA Wang and Edward~H Adelson.
\newblock Representing moving images with layers.
\newblock \emph{Transactions on Image Processing (TIP)}, 3\penalty0
  (5):\penalty0 625--638, 1994.

\bibitem[Marr and Nishihara(1978)]{marr1978representation}
David Marr and Herbert~Keith Nishihara.
\newblock Representation and recognition of the spatial organization of
  three-dimensional shapes.
\newblock \emph{Proceedings of the Royal Society of London. Series B.
  Biological Sciences}, 200\penalty0 (1140):\penalty0 269--294, 1978.

\bibitem[Binford(1971)]{binford1971visual}
I~Binford.
\newblock Visual perception by computer.
\newblock In \emph{IEEE Conference of Systems and Control}, 1971.

\bibitem[Brooks(1981)]{brooks1981symbolic}
Rodney~A Brooks.
\newblock Symbolic reasoning among 3-d models and 2-d images.
\newblock \emph{Artificial Intelligence}, 17\penalty0 (1-3):\penalty0 285--348,
  1981.

\bibitem[Kanade(1981)]{kanade1981recovery}
Takeo Kanade.
\newblock Recovery of the three-dimensional shape of an object from a single
  view.
\newblock \emph{Artificial intelligence}, 17\penalty0 (1-3):\penalty0 409--460,
  1981.

\bibitem[Broadbent(1985)]{broadbent1985question}
Donald Broadbent.
\newblock \emph{A question of levels: Comment on McClelland and Rumelhart.}
\newblock American Psychological Association, 1985.

\bibitem[Wertheimer(1912)]{wertheimer1912experimentelle}
Max Wertheimer.
\newblock Experimentelle studien uber das sehen von bewegung [experimental
  studies on the seeing of motion].
\newblock \emph{Zeitschrift fur Psychologie}, 61:\penalty0 161--265, 1912.

\bibitem[Wagemans et~al.(2012{\natexlab{a}})Wagemans, Elder, Kubovy, Palmer,
  Peterson, Singh, and von~der Heydt]{wagemans2012century}
Johan Wagemans, James~H Elder, Michael Kubovy, Stephen~E Palmer, Mary~A
  Peterson, Manish Singh, and R{\"u}diger von~der Heydt.
\newblock A century of gestalt psychology in visual perception: I. perceptual
  grouping and figure--ground organization.
\newblock \emph{Psychological bulletin}, 138\penalty0 (6):\penalty0 1172,
  2012{\natexlab{a}}.

\bibitem[Wagemans et~al.(2012{\natexlab{b}})Wagemans, Feldman, Gepshtein,
  Kimchi, Pomerantz, Van~der Helm, and Van~Leeuwen]{wagemans2012century2}
Johan Wagemans, Jacob Feldman, Sergei Gepshtein, Ruth Kimchi, James~R
  Pomerantz, Peter~A Van~der Helm, and Cees Van~Leeuwen.
\newblock A century of gestalt psychology in visual perception: Ii. conceptual
  and theoretical foundations.
\newblock \emph{Psychological bulletin}, 138\penalty0 (6):\penalty0 1218,
  2012{\natexlab{b}}.

\bibitem[K{\"o}hler(1920)]{kohler1920physischen}
Wolfgang K{\"o}hler.
\newblock \emph{Die physischen Gestalten in Ruhe und im station{\"a}renZustand.
  Eine natur-philosophische Untersuchung [The physical Gestalten at rest and in
  steady state]}.
\newblock Braunschweig, Germany: Vieweg und Sohn., 1920.

\bibitem[K{\"o}hler(1938)]{kohler1938physical}
Wolfgang K{\"o}hler.
\newblock Physical gestalten.
\newblock In \emph{A source book of Gestalt psychology}, pages 17--54. London,
  England: Routledge \& Kegan Paul, 1938.

\bibitem[Wertheimer(1923)]{wertheimer1923untersuchungen}
Max Wertheimer.
\newblock Untersuchungen zur lehre von der gestalt, ii. [investigations in
  gestalt theory: Ii. laws of organization in perceptual forms].
\newblock \emph{Psychologische Forschung}, 4:\penalty0 301--350, 1923.

\bibitem[Wertheimer(1938)]{wertheimer1938laws}
Max Wertheimer.
\newblock Laws of organization in perceptual forms.
\newblock In \emph{A source book of Gestalt psychology}, pages 71--94. London,
  England: Routledge \& Kegan Paul, 1938.

\bibitem[Koffka(2013)]{koffka2013principles}
Kurt Koffka.
\newblock \emph{Principles of Gestalt psychology}.
\newblock Routledge, 2013.

\bibitem[Lowe(2012)]{lowe2012perceptual}
David Lowe.
\newblock \emph{Perceptual organization and visual recognition}, volume~5.
\newblock Springer Science \& Business Media, 2012.

\bibitem[Pentland(1987)]{pentland1987perceptual}
Alex~P Pentland.
\newblock Perceptual organization and the representation of natural form.
\newblock In \emph{Readings in Computer Vision}, pages 680--699. Elsevier,
  1987.

\bibitem[Waltz(1975)]{waltz1975understanding}
David Waltz.
\newblock Understanding line drawings of scenes with shadows.
\newblock In \emph{The psychology of computer vision}, 1975.

\bibitem[Barrow and Tenenbaum(1981)]{barrow1981interpreting}
Harry~G Barrow and Jay~M Tenenbaum.
\newblock Interpreting line drawings as three-dimensional surfaces.
\newblock \emph{Artificial Intelligence}, 17\penalty0 (1-3):\penalty0 75--116,
  1981.

\bibitem[Lowe(1987)]{lowe1987three}
David~G Lowe.
\newblock Three-dimensional object recognition from single two-dimensional
  images.
\newblock \emph{Artificial Intelligence}, 31\penalty0 (3):\penalty0 355--395,
  1987.

\bibitem[Lowe(2004)]{lowe2004distinctive}
David~G Lowe.
\newblock Distinctive image features from scale-invariant keypoints.
\newblock \emph{International journal of computer vision}, 60\penalty0
  (2):\penalty0 91--110, 2004.

\bibitem[Coughlan and Yuille(2003)]{coughlan2003manhattan}
James~M Coughlan and Alan~L Yuille.
\newblock Manhattan world: Orientation and outlier detection by bayesian
  inference.
\newblock \emph{Neural Computation}, 2003.

\bibitem[Coughlan and Yuille(1999)]{coughlan1999manhattan}
James~M Coughlan and Alan~L Yuille.
\newblock Manhattan world: Compass direction from a single image by bayesian
  inference.
\newblock In \emph{Conference on Computer Vision and Pattern Recognition
  (CVPR)}, 1999.

\bibitem[Ren et~al.(2015)Ren, He, Girshick, and Sun]{ren2015faster}
Shaoqing Ren, Kaiming He, Ross Girshick, and Jian Sun.
\newblock Faster r-cnn: Towards real-time object detection with region proposal
  networks.
\newblock In \emph{Advances in Neural Information Processing Systems
  (NeurIPS)}, 2015.

\bibitem[He et~al.(2017)He, Gkioxari, Doll{\'a}r, and Girshick]{he2017mask}
Kaiming He, Georgia Gkioxari, Piotr Doll{\'a}r, and Ross Girshick.
\newblock Mask r-cnn.
\newblock In \emph{International Conference on Computer Vision (ICCV)}, 2017.

\bibitem[Chen et~al.(2016)Chen, Kundu, Zhang, Ma, Fidler, and
  Urtasun]{chen2016monocular}
Xiaozhi Chen, Kaustav Kundu, Ziyu Zhang, Huimin Ma, Sanja Fidler, and Raquel
  Urtasun.
\newblock Monocular 3d object detection for autonomous driving.
\newblock In \emph{Conference on Computer Vision and Pattern Recognition
  (CVPR)}, 2016.

\bibitem[Mousavian et~al.(2017)Mousavian, Anguelov, Flynn, and
  Ko{\v{s}}eck{\'a}]{mousavian20173d}
Arsalan Mousavian, Dragomir Anguelov, John Flynn, and Jana Ko{\v{s}}eck{\'a}.
\newblock 3d bounding box estimation using deep learning and geometry.
\newblock In \emph{Conference on Computer Vision and Pattern Recognition
  (CVPR)}, 2017.

\bibitem[Huang et~al.(2018{\natexlab{a}})Huang, Qi, Xiao, Zhu, Wu, and
  Zhu]{huang2018cooperative}
Siyuan Huang, Siyuan Qi, Yinxue Xiao, Yixin Zhu, Ying~Nian Wu, and Song-Chun
  Zhu.
\newblock Cooperative holistic scene understanding: Unifying 3d object, layout,
  and camera pose estimation.
\newblock In \emph{Advances in Neural Information Processing Systems
  (NeurIPS)}, 2018{\natexlab{a}}.

\bibitem[Kundu et~al.(2018)Kundu, Li, and Rehg]{kundu20183d}
Abhijit Kundu, Yin Li, and James~M Rehg.
\newblock 3d-rcnn: Instance-level 3d object reconstruction via
  render-and-compare.
\newblock In \emph{Conference on Computer Vision and Pattern Recognition
  (CVPR)}, 2018.

\bibitem[Huang et~al.(2018{\natexlab{b}})Huang, Qi, Zhu, Xiao, Xu, and
  Zhu]{huang2018holistic}
Siyuan Huang, Siyuan Qi, Yixin Zhu, Yinxue Xiao, Yuanlu Xu, and Song-Chun Zhu.
\newblock Holistic 3d scene parsing and reconstruction from a single rgb image.
\newblock In \emph{European Conference on Computer Vision (ECCV)},
  2018{\natexlab{b}}.

\bibitem[Yao et~al.(2018)Yao, Hsu, Zhu, Wu, Torralba, Freeman, and
  Tenenbaum]{yao20183d}
Shunyu Yao, Tzu~Ming Hsu, Jun-Yan Zhu, Jiajun Wu, Antonio Torralba, Bill
  Freeman, and Josh Tenenbaum.
\newblock 3d-aware scene manipulation via inverse graphics.
\newblock In \emph{Advances in Neural Information Processing Systems
  (NeurIPS)}, 2018.

\bibitem[He and Soatto(2019)]{he2019mono3d++}
Tong He and Stefano Soatto.
\newblock Mono3d++: Monocular 3d vehicle detection with two-scale 3d hypotheses
  and task priors.
\newblock \emph{arXiv preprint arXiv:1901.03446}, 2019.

\bibitem[Xiao and Furukawa(2014)]{xiao2014reconstructing}
Jianxiong Xiao and Yasutaka Furukawa.
\newblock Reconstructing the world’s museums.
\newblock \emph{International Journal of Computer Vision (IJCV)}, 2014.

\bibitem[Newell et~al.(2016)Newell, Yang, and Deng]{newell2016stacked}
Alejandro Newell, Kaiyu Yang, and Jia Deng.
\newblock Stacked hourglass networks for human pose estimation.
\newblock In \emph{European Conference on Computer Vision (ECCV)}, 2016.

\bibitem[Lee et~al.(2017)Lee, Badrinarayanan, Malisiewicz, and
  Rabinovich]{lee2017roomnet}
Chen-Yu Lee, Vijay Badrinarayanan, Tomasz Malisiewicz, and Andrew Rabinovich.
\newblock Roomnet: End-to-end room layout estimation.
\newblock In \emph{International Conference on Computer Vision (ICCV)}, 2017.

\bibitem[Zou et~al.(2018)Zou, Colburn, Shan, and Hoiem]{zou2018layoutnet}
Chuhang Zou, Alex Colburn, Qi~Shan, and Derek Hoiem.
\newblock Layoutnet: Reconstructing the 3d room layout from a single rgb image.
\newblock In \emph{Conference on Computer Vision and Pattern Recognition
  (CVPR)}, 2018.

\bibitem[Suwajanakorn et~al.(2018)Suwajanakorn, Snavely, Tompson, and
  Norouzi]{suwajanakorn2018discovery}
Supasorn Suwajanakorn, Noah Snavely, Jonathan~J Tompson, and Mohammad Norouzi.
\newblock Discovery of latent 3d keypoints via end-to-end geometric reasoning.
\newblock In \emph{Advances in Neural Information Processing Systems
  (NeurIPS)}, 2018.

\bibitem[Song et~al.(2015)Song, Lichtenberg, and Xiao]{song2015sun}
Shuran Song, Samuel~P Lichtenberg, and Jianxiong Xiao.
\newblock Sun rgb-d: A rgb-d scene understanding benchmark suite.
\newblock In \emph{Conference on Computer Vision and Pattern Recognition
  (CVPR)}, 2015.

\bibitem[Zhao and Zhu(2011)]{zhao2011image}
Yibiao Zhao and Song-Chun Zhu.
\newblock Image parsing with stochastic scene grammar.
\newblock In \emph{Advances in Neural Information Processing Systems
  (NeurIPS)}, 2011.

\bibitem[Zhao and Zhu(2013)]{zhao2013scene}
Yibiao Zhao and Song-Chun Zhu.
\newblock Scene parsing by integrating function, geometry and appearance
  models.
\newblock In \emph{Conference on Computer Vision and Pattern Recognition
  (CVPR)}, 2013.

\bibitem[Choi et~al.(2013)Choi, Chao, Pantofaru, and
  Savarese]{choi2013understanding}
Wongun Choi, Yu-Wei Chao, Caroline Pantofaru, and Silvio Savarese.
\newblock Understanding indoor scenes using 3d geometric phrases.
\newblock In \emph{Conference on Computer Vision and Pattern Recognition
  (CVPR)}, 2013.

\bibitem[Lin et~al.(2013)Lin, Fidler, and Urtasun]{lin2013holistic}
Dahua Lin, Sanja Fidler, and Raquel Urtasun.
\newblock Holistic scene understanding for 3d object detection with rgbd
  cameras.
\newblock In \emph{International Conference on Computer Vision (ICCV)}, 2013.

\bibitem[Zhang et~al.(2014)Zhang, Song, Tan, and Xiao]{zhang2014panocontext}
Yinda Zhang, Shuran Song, Ping Tan, and Jianxiong Xiao.
\newblock Panocontext: A whole-room 3d context model for panoramic scene
  understanding.
\newblock In \emph{European Conference on Computer Vision (ECCV)}, 2014.

\bibitem[Izadinia et~al.(2017)Izadinia, Shan, and Seitz]{izadinia2016im2cad}
Hamid Izadinia, Qi~Shan, and Steven~M Seitz.
\newblock Im2cad.
\newblock In \emph{Conference on Computer Vision and Pattern Recognition
  (CVPR)}, 2017.

\bibitem[Xu and Chen(2018)]{xu2018multi}
Bin Xu and Zhenzhong Chen.
\newblock Multi-level fusion based 3d object detection from monocular images.
\newblock In \emph{Conference on Computer Vision and Pattern Recognition
  (CVPR)}, 2018.

\bibitem[Chen et~al.(2015)Chen, Kundu, Zhu, Berneshawi, Ma, Fidler, and
  Urtasun]{chen20153d}
Xiaozhi Chen, Kaustav Kundu, Yukun Zhu, Andrew~G Berneshawi, Huimin Ma, Sanja
  Fidler, and Raquel Urtasun.
\newblock 3d object proposals for accurate object class detection.
\newblock In \emph{Advances in Neural Information Processing Systems
  (NeurIPS)}, 2015.

\bibitem[Straub et~al.(2014)Straub, Rosman, Freifeld, Leonard, and
  Fisher]{straub2014mixture}
Julian Straub, Guy Rosman, Oren Freifeld, John~J Leonard, and John~W Fisher.
\newblock A mixture of manhattan frames: Beyond the manhattan world.
\newblock In \emph{International Conference on Computer Vision (ICCV)}, 2014.

\bibitem[Schindler and Dellaert(2004)]{schindler2004atlanta}
Grant Schindler and Frank Dellaert.
\newblock Atlanta world: An expectation maximization framework for simultaneous
  low-level edge grouping and camera calibration in complex man-made
  environments.
\newblock In \emph{Conference on Computer Vision and Pattern Recognition
  (CVPR)}, 2004.

\bibitem[Kroeger et~al.(2015)Kroeger, Dai, and Van~Gool]{kroeger2015joint}
Till Kroeger, Dengxin Dai, and Luc Van~Gool.
\newblock Joint vanishing point extraction and tracking.
\newblock In \emph{Conference on Computer Vision and Pattern Recognition
  (CVPR)}, 2015.

\bibitem[Bosse et~al.(2003)Bosse, Rikoski, Leonard, and
  Teller]{bosse2003vanishing}
Michael Bosse, Richard Rikoski, John Leonard, and Seth Teller.
\newblock Vanishing points and three-dimensional lines from omni-directional
  video.
\newblock \emph{The Visual Computer}, 2003.

\bibitem[Straub et~al.(2015)Straub, Bhandari, Leonard, and
  Fisher]{straub2015real}
Julian Straub, Nishchal Bhandari, John~J Leonard, and John~W Fisher.
\newblock Real-time manhattan world rotation estimation in 3d.
\newblock In \emph{International Conference on Intelligent Robots and Systems
  (IROS)}, 2015.

\bibitem[Ghanem et~al.(2015)Ghanem, Thabet, Carlos~Niebles, and
  Caba~Heilbron]{ghanem2015robust}
Bernard Ghanem, Ali Thabet, Juan Carlos~Niebles, and Fabian Caba~Heilbron.
\newblock Robust manhattan frame estimation from a single rgb-d image.
\newblock In \emph{Conference on Computer Vision and Pattern Recognition
  (CVPR)}, 2015.

\bibitem[Hedau et~al.(2009)Hedau, Hoiem, and Forsyth]{hedau2009recovering}
Varsha Hedau, Derek Hoiem, and David Forsyth.
\newblock Recovering the spatial layout of cluttered rooms.
\newblock In \emph{Conference on Computer Vision and Pattern Recognition
  (CVPR)}, 2009.

\bibitem[Lee et~al.(2009)Lee, Hebert, and Kanade]{lee2009geometric}
David~C Lee, Martial Hebert, and Takeo Kanade.
\newblock Geometric reasoning for single image structure recovery.
\newblock In \emph{Conference on Computer Vision and Pattern Recognition
  (CVPR)}, 2009.

\bibitem[Hedau et~al.(2010)Hedau, Hoiem, and Forsyth]{hedau2010thinking}
Varsha Hedau, Derek Hoiem, and David Forsyth.
\newblock Thinking inside the box: Using appearance models and context based on
  room geometry.
\newblock In \emph{European Conference on Computer Vision (ECCV)}, 2010.

\bibitem[Schwing et~al.(2012)Schwing, Hazan, Pollefeys, and
  Urtasun]{schwing2012efficient}
Alexander~G Schwing, Tamir Hazan, Marc Pollefeys, and Raquel Urtasun.
\newblock Efficient structured prediction for 3d indoor scene understanding.
\newblock In \emph{Conference on Computer Vision and Pattern Recognition
  (CVPR)}, 2012.

\bibitem[Delage et~al.(2007)Delage, Lee, and Ng]{delage2007automatic}
Erick Delage, Honglak Lee, and Andrew~Y Ng.
\newblock Automatic single-image 3d reconstructions of indoor manhattan world
  scenes.
\newblock In \emph{Robotics Research}, pages 305--321. Springer, 2007.

\bibitem[Furukawa et~al.(2009)Furukawa, Curless, Seitz, and
  Szeliski]{furukawa2009manhattan}
Yasutaka Furukawa, Brian Curless, Steven~M Seitz, and Richard Szeliski.
\newblock Manhattan-world stereo.
\newblock In \emph{Conference on Computer Vision and Pattern Recognition
  (CVPR)}, 2009.

\bibitem[Xiao et~al.(2013)Xiao, Hays, Russell, Patterson, Ehinger, Torralba,
  and Oliva]{xiao2013basic}
Jianxiong Xiao, James Hays, Bryan~C Russell, Genevieve Patterson, Krista
  Ehinger, Antonio Torralba, and Aude Oliva.
\newblock Basic level scene understanding: categories, attributes and
  structures.
\newblock \emph{Frontiers in psychology}, 4:\penalty0 506, 2013.

\bibitem[Ren and Sudderth(2016)]{ren2016three}
Zhile Ren and Erik~B Sudderth.
\newblock Three-dimensional object detection and layout prediction using clouds
  of oriented gradients.
\newblock In \emph{Conference on Computer Vision and Pattern Recognition
  (CVPR)}, 2016.

\bibitem[Liu et~al.(2017)Liu, Zhao, and Zhu]{liu2017single}
Xiaobai Liu, Yibiao Zhao, and Song-Chun Zhu.
\newblock Single-view 3d scene reconstruction and parsing by attribute grammar.
\newblock \emph{Transactions on Pattern Analysis and Machine Intelligence
  (TPAMI)}, 40\penalty0 (3):\penalty0 710--725, 2017.

\bibitem[Wu et~al.(2017)Wu, Wang, Xue, Sun, Freeman, and
  Tenenbaum]{wu2017marrnet}
Jiajun Wu, Yifan Wang, Tianfan Xue, Xingyuan Sun, Bill Freeman, and Josh
  Tenenbaum.
\newblock Marrnet: 3d shape reconstruction via 2.5 d sketches.
\newblock In \emph{Advances in Neural Information Processing Systems
  (NeurIPS)}, 2017.

\bibitem[Zhu et~al.(2018)Zhu, Zhang, Zhang, Wu, Torralba, Tenenbaum, and
  Freeman]{von2018}
Jun-Yan Zhu, Zhoutong Zhang, Chengkai Zhang, Jiajun Wu, Antonio Torralba,
  Joshua~B. Tenenbaum, and William~T. Freeman.
\newblock Visual object networks: Image generation with disentangled 3d
  representations.
\newblock In \emph{Advances in Neural Information Processing Systems
  (NeurIPS)}, 2018.

\bibitem[Zhang et~al.(2018)Zhang, Zhang, Zhang, Tenenbaum, Freeman, and
  Wu]{genre2018}
Xiuming Zhang, Zhoutong Zhang, Chengkai Zhang, Joshua~B Tenenbaum, William~T
  Freeman, and Jiajun Wu.
\newblock Learning to reconstruct shapes from unseen classes.
\newblock In \emph{Advances in Neural Information Processing Systems
  (NeurIPS)}, 2018.

\bibitem[Tulsiani et~al.(2018)Tulsiani, Gupta, Fouhey, Efros, and
  Malik]{tulsiani2017factoring}
Shubham Tulsiani, Saurabh Gupta, David Fouhey, Alexei~A Efros, and Jitendra
  Malik.
\newblock Factoring shape, pose, and layout from the 2d image of a 3d scene.
\newblock In \emph{Conference on Computer Vision and Pattern Recognition
  (CVPR)}, 2018.

\bibitem[Wu et~al.(2016)Wu, Xue, Lim, Tian, Tenenbaum, Torralba, and
  Freeman]{wu2016single}
Jiajun Wu, Tianfan Xue, Joseph~J Lim, Yuandong Tian, Joshua~B Tenenbaum,
  Antonio Torralba, and William~T Freeman.
\newblock Single image 3d interpreter network.
\newblock In \emph{European Conference on Computer Vision (ECCV)}, 2016.

\bibitem[Tekin et~al.(2018)Tekin, Sinha, and Fua]{tekin2018real}
Bugra Tekin, Sudipta~N Sinha, and Pascal Fua.
\newblock Real-time seamless single shot 6d object pose prediction.
\newblock In \emph{Conference on Computer Vision and Pattern Recognition
  (CVPR)}, 2018.

\bibitem[Girshick(2015)]{girshick2015fast}
Ross Girshick.
\newblock Fast r-cnn.
\newblock In \emph{International Conference on Computer Vision (ICCV)}, 2015.

\bibitem[Olshausen and Field(1996)]{olshausen1996emergence}
Bruno~A Olshausen and David~J Field.
\newblock Emergence of simple-cell receptive field properties by learning a
  sparse code for natural images.
\newblock \emph{Nature}, 381\penalty0 (6583):\penalty0 607, 1996.

\bibitem[Wu et~al.(2010)Wu, Si, Gong, and Zhu]{wu2010learning}
Ying~Nian Wu, Zhangzhang Si, Haifeng Gong, and Song-Chun Zhu.
\newblock Learning active basis model for object detection and recognition.
\newblock \emph{International Journal of Computer Vision (IJCV)}, 90\penalty0
  (2):\penalty0 198--235, 2010.

\bibitem[Rezende et~al.(2016)Rezende, Eslami, Mohamed, Battaglia, Jaderberg,
  and Heess]{rezende2016unsupervised}
Danilo~Jimenez Rezende, SM~Ali Eslami, Shakir Mohamed, Peter Battaglia, Max
  Jaderberg, and Nicolas Heess.
\newblock Unsupervised learning of 3d structure from images.
\newblock In \emph{Advances in Neural Information Processing Systems
  (NeurIPS)}, 2016.

\bibitem[Yan et~al.(2016)Yan, Yang, Yumer, Guo, and Lee]{yan2016perspective}
Xinchen Yan, Jimei Yang, Ersin Yumer, Yijie Guo, and Honglak Lee.
\newblock Perspective transformer nets: Learning single-view 3d object
  reconstruction without 3d supervision.
\newblock In \emph{Advances in Neural Information Processing Systems
  (NeurIPS)}, 2016.

\bibitem[He et~al.(2016)He, Zhang, Ren, and Sun]{he2016deep}
Kaiming He, Xiangyu Zhang, Shaoqing Ren, and Jian Sun.
\newblock Deep residual learning for image recognition.
\newblock In \emph{Conference on Computer Vision and Pattern Recognition
  (CVPR)}, 2016.

\bibitem[Lin et~al.(2017)Lin, Doll{\'a}r, Girshick, He, Hariharan, and
  Belongie]{lin2017feature}
Tsung-Yi Lin, Piotr Doll{\'a}r, Ross Girshick, Kaiming He, Bharath Hariharan,
  and Serge Belongie.
\newblock Feature pyramid networks for object detection.
\newblock In \emph{Conference on Computer Vision and Pattern Recognition
  (CVPR)}, 2017.

\bibitem[Massa and Girshick(2018)]{massa2018mrcnn}
Francisco Massa and Ross Girshick.
\newblock {maskrcnn-benchmark: Fast, modular reference implementation of
  Instance Segmentation and Object Detection algorithms in PyTorch}.
\newblock \url{https://github.com/facebookresearch/maskrcnn-benchmark}, 2018.

\bibitem[Geiger et~al.(2013)Geiger, Lenz, Stiller, and
  Urtasun]{geiger2013vision}
Andreas Geiger, Philip Lenz, Christoph Stiller, and Raquel Urtasun.
\newblock Vision meets robotics: The kitti dataset.
\newblock \emph{International Journal of Robotics Research (IJRR)}, 32\penalty0
  (11):\penalty0 1231--1237, 2013.

\end{thebibliography}

\end{document}